\documentclass[preprint,12pt]{elsarticle}
\usepackage{rotating}
\usepackage{graphics}
\usepackage{multirow}
\usepackage{hyperref}
\usepackage{caption}
\usepackage{subcaption}
\usepackage{url}
\usepackage{algorithm}
\usepackage{algorithmic}
\usepackage{ulem}
\usepackage{xcolor,tikz}
\usepackage{soul}
\usepackage{booktabs}
\usepackage{setspace}
\floatname{algorithm}{Block}

\newcommand\reduline{%
 \bgroup\markoverwith
  {\textcolor{red}{\pgfsetfillopacity{0.2}\rule[-0.5ex]{2pt}{10pt}\pgfsetfillopacity{1}}%
   \textcolor{red}{\llap{\rule[0.4ex]{2pt}{0.4pt}}\llap{\rule[0.7ex]{2pt}{0.4pt}}}%
  }%
  \ULon}
\journal{Journal of Biomedical Informatics}

\begin{document}
\begin{frontmatter}

\title{Useful Blunders: Can Automated Speech Recognition Errors Improve Downstream Dementia Classification?}

\author[inst1]{Changye Li}

\affiliation[inst1]{organization={Institute of Health Informatics},
            addressline={University of Minnesota}, 
            city={Minneapolis},
            postcode={55455}, 
            state={Minnesota},
            country={USA}}

\author[inst2]{Weizhe Xu}

\affiliation[inst2]{organization={Biomedical Informatics and Medical Education},
            addressline={University of Washington}, 
            city={Seattle},
            postcode={98195}, 
            state={Washington},
            country={USA}}
            
\author[inst2]{Trevor Cohen}

\author[inst3]{Serguei Pakhomov}

\affiliation[inst3]{organization={Department of Pharmaceutical Care \& Health Systems},
            addressline={University of Minnesota}, 
            city={Minneapolis},
            postcode={55455}, 
            state={Minnesota},
            country={USA}}

\begin{abstract}

\textbf{Objectives}: We aimed to investigate how errors from automatic speech recognition (ASR) systems affect dementia classification accuracy, specifically in the ``Cookie Theft'' picture description task. We aimed to assess whether imperfect ASR-generated transcripts could provide valuable information for distinguishing between language samples from cognitively healthy individuals and those with Alzheimer's disease (AD).

\textbf{Methods}: We conducted experiments using various ASR models, refining their transcripts with post-editing techniques. Both these imperfect ASR transcripts and manually transcribed ones were used as inputs for the downstream dementia classification. We conducted comprehensive error analysis to compare model performance and assess ASR-generated transcript effectiveness in dementia classification.

\textbf{Results}: Imperfect ASR-generated transcripts surprisingly outperformed manual transcription for distinguishing between individuals with AD and those without in the ``Cookie Theft'' task. These ASR-based models surpassed the previous state-of-the-art approach, indicating that ASR errors may contain valuable cues related to dementia. The synergy between ASR and classification models improved overall accuracy in dementia classification.

\textbf{Conclusion}: Imperfect ASR transcripts effectively capture linguistic anomalies linked to dementia, improving accuracy in classification tasks. This synergy between ASR and classification models underscores ASR's potential as a valuable tool in assessing cognitive impairment and related clinical applications.

\end{abstract}

\begin{highlights}
\item Automatic speech recognition (ASR) errors can provide important information for downstream dementia classification, resulting in better performance than with professional manual transcripts.

\item Our comprehensive error analysis provides new insights into linguistic anomalies in dementia that inform text classification with transformer models.
\end{highlights}

\end{frontmatter}

\section{Introduction}

Alzheimer's disease (AD) is a neurodegenerative disorder that affects the use of speech and language and is difficult to diagnose in its early stages. As a result, the number of people living with undiagnosed AD is continuously growing \citep{gaugler20222022}. Absence or delayed diagnosis can result in unnecessary negative effects on dementia patients and their caregivers \citep{https://doi.org/10.1111/psyg.12095}. To address this growing problem, simple, timely, and cost-effective diagnostic strategies are urgently needed \citep{fox2013pros, iliffe2003sooner}. 

Current methods of AD diagnosis include multiple sources of information such as caregiver reports, a consensus agreement among the examining nurse, physician, and neuropsychologist, imaging results, structured interviews, and cognitive tests that include examination of language use characteristics. One of the disadvantages of comprehensive cognitive testing for AD is that test batteries are lengthy, conducted in a controlled laboratory setting, and may require longitudinal observation and monitoring before reaching the final diagnosis. Such tests can sometimes be insensitive to naturalistic language patterns \citep{sabat1994language} and early signs of the linguistic deficits caused by AD in daily communication \citep{crockford1994assessing}. Spontaneous speech has proven to be a valuable source of information helpful to assess an individual's cognitive state. The assessment of spontaneous speech is sensitive to the linguistic anomalies caused by AD, providing earlier and more accurate diagnosis \citep{bucks2000analysis}. However, manually examining the transcripts of structured interviews, cognitive/neurological tests and spontaneous speech can be labor- and time-intensive.  

To automate the assessment of linguistic anomalies, the machine learning community developed multiple approaches for analysis of language and speech including natural language processing (NLP) methods based on supervised learning, an approach that learns patterns of associations between inputs and outputs from labeled data (for a review, see \citet{martinez2021ten}). In recent years, transfer learning with pre-trained neural language models (NLMs) has become increasingly popular and has met with success on various downstream tasks \citep{gruetzemacher2022deep}. Transfer learning leverages the knowledge from pre-trained NLMs and applies it to a different, and typically under-resourced, task without the requirement of training large new models from scratch. For AD classification, prior work shows that fine-tuning pre-trained NLMs to even very small datasets can outperform machine learning approaches that rely on hand-crafted features \citep{balagopalan20_interspeech}.

To realize their benefits for assessment of cognitive impairment, NLP models require verbatim transcriptions of speech from patients. This presents a bottleneck for the acquisition of data in research, and in practice. Automatic speech recognition (ASR) models can be used to generate transcripts from audio recordings automatically, and the generated transcripts have shown to be a good source of features for the downstream classification task of AD detection \citep{weiner2017manual}. A major benefit of this ASR-driven approach is that it can eliminate the bottleneck of manual transcription, a process that introduces resource and privacy constraints. Using ASR promises to provide a path toward large-scale deployment of linguistically-informed dementia screening methods. However, ASR errors (typically measured with word error rate (WER) and character error rate (CER)) have been previously shown to impede predictive models' performance in identifying dementia from audio samples by lowering their classification accuracy \citep{zhou16_interspeech}. 

ASR performance has improved considerably in recent years, with an important inflexion point being the development of self-supervised learning (SSL) ASR models such as Wav2Vec2 \citep{NEURIPS2020_92d1e1eb} and HuBERT \citep{ Hsu2021HuBERTSS}. SSL is a subcategory of unsupervised learning that concerns approaches to extract information from the input data itself as the source of labels used to learn internal representations (for reviews, see \citet{mohamed2022self, ericsson2022self}), thus alleviating the manual annotation bottleneck. These models can achieve single digit WERs and CERs on open domain datasets\footnote{For the benchmark statistics, see:\url{https://paperswithcode.com/sota/speech-recognition-on-librispeech-test-clean}}. However, this high level of performance does not directly translate to clinical or related settings. Previous studies have reported much higher WERs ($\approx$ 40\%) for ASR models on speech collected in clinically-related contexts \citep{XU2022103998, min-etal-2021-evaluating, Sadeghian2021, DBLP:journals/corr/abs-2110-15704}. Furthermore, there has been scant work investigating ASR performance on the speech of dementia patients and the utility of ASR-generated transcripts for downstream dementia classification \citep{balagopalan-etal-2020-impact, Sadeghian2021, DBLP:journals/corr/abs-2110-15704}.

In a previous recent study \citep{li2022far}, we conducted a preliminary investigation of the impact of ASR errors found in the transcripts of picture descriptions by dementia patients and controls on classification performance of pre-trained NLMs. We found a non-linear relationship, showing that ASR errors do not necessarily lead to decreased classification performance. However, in that study we used ASR-generated transcripts only for the \textit{testing} data used in our experiments with ASR models. On the training side, we used ``perfect'' manually created transcripts for the data that were used to fine-tune NLMs used to categorize samples. This previous study raised the question of whether using the same transcription method at training and test time might lead to better overall performance.

In the current study, we investigate in depth not only the impact of errors present in imperfect \textit{test} transcripts produced by ASR models on their subsequent categorization, but also the effect of ASR errors in the language samples used for fine-tuning in the first place. Our overarching hypothesis is that speech deficits in patients with dementia may lead to systematic ASR errors that a categorization model can leverage to improve its classification performance further. In the remainder of this paper, we demonstrate that this is indeed the case; however, there are several important nuances that we will discuss in detail further. The contributions of this study can be summarized as follows: a) we leverage post-editing methods to enhance the performance of ASR models, approaching the previously reported evaluation metrics on challenging spontaneous speech audio recordings produced by dementia patients and healthy individuals; b) our investigation shows speech deficits in patients with dementia lead to systematic ASR errors that can be harnessed to enhance classification performance; c) our study does not only aim to pinpoint specific ASR models for dementia classification but also provides insights into the characteristics and mechanisms that would make ASR models more beneficial for such tasks; d) our comprehensive analysis lays the foundation for guiding the future development of ASR models and workflows for detecting cognitive impairment, aiming to collectively improve their performance and applicability in dementia screening and similar clinical applications. The code developed for this work and metadata and tools necessary to reproduce our findings are publicly available on GitHub\footnote{\url{https://github.com/LinguisticAnomalies/paradox-asr}}.

\begin{table}[ht!]
\fontsize{9pt}{9pt}
\resizebox{\textwidth}{!}{%
\begin{tabular}{@{}|p{0.3\linewidth}|p{0.35\linewidth}|p{0.35\linewidth}|@{}}
\toprule
\textbf{Problem or Issue}  & \textbf{What is Already Known} & \textbf{What this Paper Adds} \\ \midrule
Efforts to develop NLP models for detecting linguistic anomalies associated with dementia face the challenge of acquiring verbatim transcripts of speech from patients, which often necessitates time-consuming manual transcription. Additionally, the performance of ASR systems is known to influence the accuracy of downstream classification tasks, raising the need to understand the implications of ASR errors in this context.& Current ASR systems, despite notable improvements, continue to exhibit errors, particularly when confronted with challenging audio data. Previous research has already established that the performance of ASR systems is a critical factor in various applications.  & This study explores the applicability of ASR-generated transcripts in the context of dementia classification. It reveals that ASR errors can contribute valuable features, and the utilization of imperfect ASR transcripts surpasses the performance of manually generated transcripts. This highlights a remarkable synergy between ASR and classification models in identifying dementia manifestations, enriching our understanding of the potential of ASR technology in clinical applications. \\ \bottomrule
\end{tabular}}
\caption*{Statement of significance}
\label{tab:my-table}
\end{table}
\pagebreak

\section{Methods}

\subsection{Data}
\begin{figure}[htbp]
    \centering
    \includegraphics[width=0.6\textwidth]{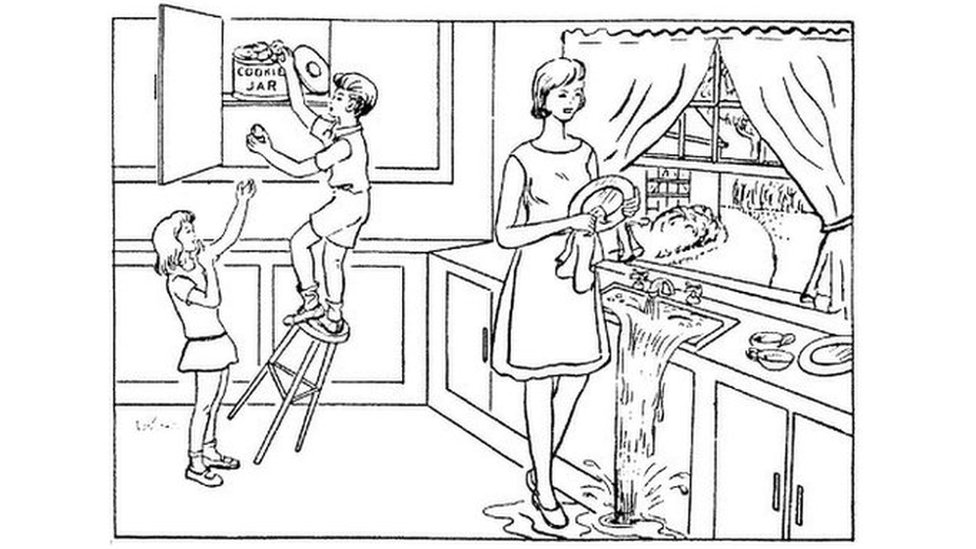}
    \caption{The ``Cookie Theft'' picture description stimuli. In this task, participants are presented with this picture stimuli and are asked to describe everything they observe in the picture.}
    \label{fig:cookie-theft}
\end{figure}

We used two publicly available datasets: a) Alzheimer's Dementia (AD) Recognition through Spontaneous Speech (ADReSS) \citep{bib:LuzHaiderEtAl20ADReSS}, and b) recordings and transcripts drawn from the Wisconsin Longitudinal Study (WLS) \citep{herd2014cohort}. Both datasets contain audio recordings of participants in a widely-used diagnostic task, the ``Cookie Theft'' picture description (Figure~\ref{fig:cookie-theft}) from the Boston Diagnostic Aphasia Examination \citep{goodglass1983boston}. This task has shown to reveal subtle semantic deficits that may emerge even in the initial stages of the illness, as demonstrated in spontaneous speech \citep{forbes2005detecting}. In this task, participants were asked to describe everything they see going on in the picture, and their responses were audio recorded. The audio samples in both datasets were manually transcribed verbatim and annotated in CHAT \citep{10.1162/coli.2000.26.4.657} format with the Computerized Language Analysis (CLAN) annotation system.

In each dataset, we performed basic pre-processing of both verbatim transcripts and audio recordings using TRESTLE (\textbf{T}oolkit for \textbf{R}eproducible \textbf{E}xecution of \textbf{S}peech \textbf{T}ext and \textbf{L}anguage \textbf{E}xperiments)\footnote{\url{https://github.com/LinguisticAnomalies/harmonized-toolkit}} \citep{li2023trestle} by a) removing artifacts such as speech and non-speech event descriptions (i.e., ``overlap'', ``clear throat''), unintelligible words, b) speech/transcripts that did not belong to participants, c) resampling the audio recordings to 16 kHz to match the sampling rate of the audio used to train ASR models, and d) partitioning the audio recordings into utterance-level chunks using the time-stamps provided in the CHAT-formatted transcripts.

The ADReSS dataset is a subset of the DementiaBank English Pitt Corpus \footnote{https://dementia.talkbank.org/access/English/Pitt.html}\cite{becker1994natural}. In addition to the picture description task, all participants contributing data to this corpus underwent a series of neurocognitive tests including the Mini Mental State Examination (MMSE) \citep{folstein1975mini}. The ADReSS dataset contains a total of 156 samples from 78 healthy controls and 78 dementia patients matched on age and sex \citep{bib:LuzHaiderEtAl20ADReSS}. Automated transcriptions of the audio recordings were used to train and evaluate models for dementia classification. Using TRESTLE, we generated an utterance-level corpus split into training and test portions with 1,414 and 567 utterance-level transcripts in the training and test splits, respectively. 

The WLS is an ongoing longitudinal study of a random sample of 10,317 men and women who graduated from Wisconsin high schools in 1957. The participants were interviewed up to six times between 1957 and 2011. All of the participants in the WLS were considered to be cognitively healthy at the start of the study, however, some of them may have developed dementia and other forms of cognitive impairment in the later years, as would be expected based on incidence of these conditions in the general population. The ``Cookie Theft'' picture description task was introduced in the later rounds of interviews in 2011. While all of the audio of the picture description task has been transcribed verbatim, only a portion of the transcripts have been manually verified and aligned/linked with the audio. These aligned transcripts were used to train ASR models, and evaluate their accuracy. For the current study, we restricted the original WLS dataset to 87 female and 100 male participants, who a) agreed to participate in the ``Cookie Theft'' picture description task in 2011 and b) had recording samples with manually aligned verbatim transcripts. We selected the first 52 female and 62 male participants and their output as the training set, and the remaining participants as the evaluation set, resulting in 1,701 and 1,146 utterance-level transcripts, respectively. We used the professionally transcribed transcripts that have not yet been verified and linked with the audio from the remaining 1,182 participants to build a 5-gram language model subsequently used in transcript generation. Key characteristics of the ADReSS and WLS datasets used in this study are provided in Table~\ref{tab:my-data}. 

As is described in further detail in the next section, we used the ADReSS dataset to assess classification performance for detection of dementia. The WLS dataset was used only to domain-adapt pre-trained ASR models and to train a language model for ASR systems that included the Connectionist Temporal Classification (CTC) \citep{10.1145/1143844.1143891} decoding of audio samples\footnote{Researchers interested in comparing their results to ours could use the TRESTLE package along with a machine-readable manifest file provided upon request to recreate our exact experimental environment}.

\begin{table}[htbp]
\centering
\resizebox{\textwidth}{!}{
\begin{tabular}{|ll|ll|lll|}
\hline
\multicolumn{2}{|l|}{\multirow{2}{*}{\textbf{Characteristics}}} & \multicolumn{2}{l|}{\textbf{ADReSS}}    & \multicolumn{3}{l|}{\textbf{WLS}}                            \\ \cline{3-7} 
\multicolumn{2}{|l|}{}& \multicolumn{1}{l|}{Train} & Test & \multicolumn{1}{l|}{Train} & \multicolumn{1}{l|}{Test} & Remaining \\ \hline
\multicolumn{2}{|l|}{Age, mean (SD)}                  & \multicolumn{1}{l|}{65.6 (6.5)} & 65.6 (7.0) & \multicolumn{1}{l|}{70.2 (3.2)} & \multicolumn{1}{l|}{70.3 (3.6)} &  70.4 (4.6)\\ \hline
\multicolumn{1}{|l|}{\multirow{2}{*}{Gender, n (\%)}} & Male & \multicolumn{1}{l|}{48 (44)} & 22 (46) & \multicolumn{1}{l|}{62 (53)} & \multicolumn{1}{l|}{38 (48)} & 594 (50) \\ \cline{2-7} 
\multicolumn{1}{|l|}{}  & Female & \multicolumn{1}{l|}{60 (56)} & 26 (54) & \multicolumn{1}{l|}{52 (47)} & \multicolumn{1}{l|}{35 (48)} & 588 (50) \\ \hline
\multicolumn{2}{|l|}{Education, mean (SD)} & \multicolumn{1}{l|}{13.1 (3)} &12.9 (2.0)  & \multicolumn{1}{l|}{13.3 (2.1)} & \multicolumn{1}{l|}{13.1 (3.7)} & 13.6 (3.1) \\ \hline
\multicolumn{2}{|l|}{MMSE, mean (SD)} & \multicolumn{1}{l|}{24.1 (5.9)} &24.9 (5.3) & \multicolumn{1}{l|}{NA} & \multicolumn{1}{l|}{NA} & NA \\ \hline
\multicolumn{2}{|l|}{Number of utterance-level transcripts}  & \multicolumn{1}{l|}{1,414} & 567 & \multicolumn{1}{l|}{1,701} & \multicolumn{1}{l|}{1,146} & 16,043 \\ \hline
\end{tabular}
}
\caption{Participant characteristics. Age of the WLS participants was taken from the 2011 visit when the ``Cookie Theft'' test was administered.}
\label{tab:my-data}
\end{table}

\subsection{Models}

In this study, we investigated the performance of two pre-trained ASR models: Wav2Vec2 and HuBERT, and experimented with using a Bidirectional Encoder Representations from Transformer (BERT) model for classification of ASR-generated transcripts. All three of these models are based on the Transformer neural network architecture \citep{NIPS2017_3f5ee243}.

\subsubsection{Wav2Vec2}

Wav2Vec2 is a Transformer neural network built on top of a convolutional neural network (CNN) \citep{lecun1989handwritten}. With speech, CNNs learn to extract ASR-relevant features from audio waveforms.  During the pre-training stage, Wav2Vec2 consumes input waveforms and extracts the resulting convolutional features. Wav2Vec2 generates hidden units for the encoder layers of a Transformer neural model \citep{NIPS2017_3f5ee243} using contrastive predictive coding (CPC) \citep{oord2018representation}. CPC is a technique that learns self-supervised representations by predicting the next element in the sequence given the elements that precede it (analogous to the training objective of generative language models such as OpenAI's GPT, but predicting further than one step ahead in the sequence and contrasting observations with randomly-drawn counterexamples). Learning is achieved by maximizing the probabilistic contrastive loss \citep{NEURIPS2020_d89a66c7} to capture maximally useful information for future prediction. 

\subsubsection{HuBERT}

In contrast to Wav2vec2, HuBERT was inspired by masked language modeling, the training objective for Bidirectional Encoder Representations from Transformer (BERT) \citep{devlin-etal-2019-bert}. Similarly to BERT, HuBERT masks input sequences and then attempts to reconstruct them, a pre-training process that can be conducted without the need for corresponding transcripts (the unmasked sequences provide the ``self'' supervision). With BERT, the masked features are sequences of tokens representing words, and/or their character sub-sequences. In constrast, HuBERT consumes continuous input sequences consisting of frames composed of Mel-frequency cepstral coefficients (MFCCs)\footnote{MFCC features are derived from the raw audio signal in a sequence of transformations that emphasize higher frequencies (``pre-emphasis'') using the discrete Fourier transform with subsequent numerical transformations to Cepstral coefficients on the Mel scale that represents the range and granularity of human sound perception.} and trains on classic k-means units for feature extraction. HuBERT uses pseudo-phonetic labels for each frame to predict the masked and pre-computed k-means clusters of MFCC features with cross-entropy loss over two rounds of pre-training. In the first round, the latent HuBERT features and pre-computed k-means clusters provide the target vocabulary. In the second round, HuBERT trains to predict the targets at masked positions using the target vocabulary with the cross-entropy loss. This strategy enables HuBERT to learn a combined acoustic and language model over continuous inputs.

\subsubsection{BERT}

As with the Wav2vec2 and HuBERT ASR models, the BERT language model is pre-trained in a self-supervised manner, in this case using a masked language modeling objective. Pre-trained BERT models exemplify the power of transfer learning by leveraging information acquired during the pre-training to improve performance on downstream tasks with relatively small amounts of domain-specific training data needed for fine-tuning the pre-trained model. During the fine-tuning step, the general distribution of tokens, their relative positions, and model weights learned during the pre-training can be either directly applied, or applied with minimal changes, to the downstream task. 

In \citet{balagopalan20_interspeech}, the authors report fine-tuning a BERT model on the ADReSS training set. The fine-tuned BERT model achieved an accuracy of 0.826 and area under the receiver-operator curve (AUC) of 0.873 on the ADReSS test set, the best results reported on this set for models trained on text alone. In the current study, we first merged the utterance-level ASR-generated transcripts into participant-level transcripts for both the ADReSS training and test sets. We then fine-tuned the BERT base model with a sequential classification head with a batch size of 8 over 10 epochs using ASR-generated transcripts from the ADReSS training set. On the ADReSS test set, we evaluated the fine-tuned classification model performance using the ASR-generated transcripts with accuracy and AUC metrics.

\subsubsection{Model Variants}

While pre-training of all Wav2Vec2 and HuBERT model variants involved the 960 hour portion of the audio from the LibriSpeech dataset \citep{7178964} \textit{without} using the corresponding transcripts, the model variants differed in terms of additional data and methods used in pre-training and subsequent adaptation for general-purpose ASR. For example, {wav2vec2-base-960h}\footnote{\url{https://huggingface.co/facebook/wav2vec2-base-960h}}, {wav2vec2-large-960h}\footnote{\url{https://huggingface.co/facebook/wav2vec2-large-960h}}, and {hubert-lage-ls-960-ft}\footnote{\url{https://huggingface.co/facebook/hubert-large-ls960-ft}} were pre-trained and subsequently domain-adapted for general-purpose ASR on the same 960 hours portion of the LibriSpeech corpus. In contrast, {wav2vec2-large-960h-lv60}\footnote{\url{https://huggingface.co/facebook/wav2vec2-large-960h-lv60}} was pre-trained and subsequently adapted for general-purpose ASR using the LibriSpeech and LibriLight \citep{librilight} datasets, whereas {wav2vec2-large-960h-lv60-self}\footnote{\url{https://huggingface.co/facebook/wav2vec2-large-960h-lv60-self}} was pre-trained and adapted using the same dataset as with {wav2vec2-large-960h} but with a self-training objective \citep{Xu2021SelfTrainingAP}. The self-training objective here represents semi-supervised pseudo-label-style self-training \citep{9054295, DBLP:journals/corr/abs-1911-08460}. In self-training, the acoustic model (i.e., Wav2Vec2) is pre-trained on available unlabeled  data (i.e., manual transcripts are not present) to an ``intermediate'' acoustic model, then task-adapted on the available labeled data (i.e., manual transcripts are present). The ``intermediate'' model then pseudo-labels another unlabeled dataset for additional training data and is trained as the ``final'' acoustic model.

To emphasize the distinction between three different types of fine-tuning of the models employed in the current work, we define three types of adaptation on speech and/or text data as follows: 

\begin{enumerate}
    \item Using pre-trained Wav2Vec2 and HuBERT models without modification. We refer to these ASR models as \textit{pre-trained ASR} models.
    \item Fine-tuning Wav2Vec2 and HuBERT models with additional ``Cookie Theft'' picture description from the WLS training set.  We refer to these ASR models as \textit{domain-adapted ASR} models
    \item Fine-tuning pre-trained BERT models to classify ASR-generated transcripts into dementia vs. controls categories. We refer to these BERT models as \textit{fine-tuned classification} models.
\end{enumerate}

\subsubsection{CTC and ASR Decoding Methods}
\label{section:CTC}
All ASR models have a CTC \citep{10.1145/1143844.1143891} layer at the top of their architectures and were paired with adapted post-processors that decode features extracted from input frames and model outputs to generate word-level transcripts. 

CTC calculates the loss between continuous unsegmented input sequences and target sequences by summing over the probability of possible alignments between inputs and outputs, providing label sequences and alignment simultaneously during model training. Specifically, CTC encodes the audio input and assigns probabilities to a character-level matrix that represents a lattice of characters hypothesized by the model at each time step during the decoding. By default, CTC generates the output using the best-path decoding method. Best-path decoding consists of calculating the loss by summing up all possible alignment scores via the character-level matrix and generates the output by computing a path through the lattice of matrix cells containing the character with the highest probability at each time step.

Selecting only one most likely character at each time step does not \textit{always} guarantee the optimal sequence. To address this issue, beam search may be used to create a beam of $k$ (the beam size) best paths during each decoding step. At each time step, the search chooses $k$ characters with the highest conditional probability until the end of the time step. In addition to the beam search, CTC employs byte pair encoding (BPE) \citep{sennrich-etal-2016-neural} to select the best $k$ characters at each time step. BPE alleviates the out-of-vocabulary problem by tokenizing less frequent or unknown words as sequences of smaller subwords and encoding the more common words as single tokens. Using these methods, we trained an n-gram language model to find the best sequence amongst these $k$ possible outputs. Subsequently, the highest weighted sum of two possibilities (i.e., the best ones from the beam search decoding method and the language model, respectively) is chosen as the final output.

\subsection{Evaluation}
\begin{figure}
\begin{subfigure}{\textwidth}
  \centering
  \includegraphics[width=\linewidth]{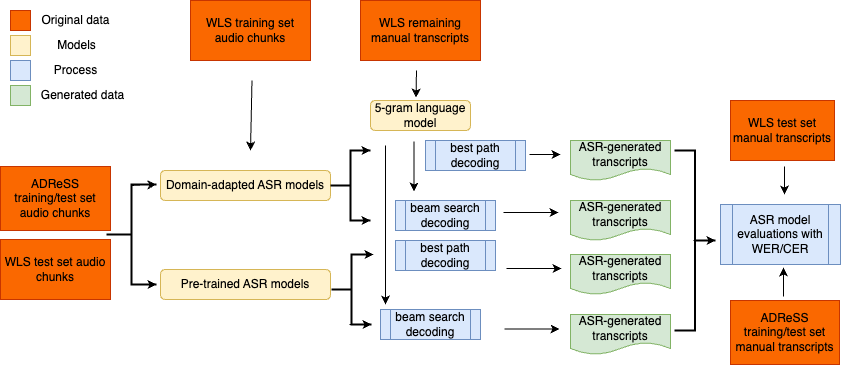}  
  \caption{Phase 1: generating transcripts from utterance-level audio segments with pre-trained and adapted ASR models. ASR models are evaluated using WER/CER between ASR-generated transcripts and ``prefect'' manual transcripts.}
  \label{fig:phase1}
\end{subfigure}
\begin{subfigure}{\textwidth}
  \centering
  \includegraphics[width=\linewidth]{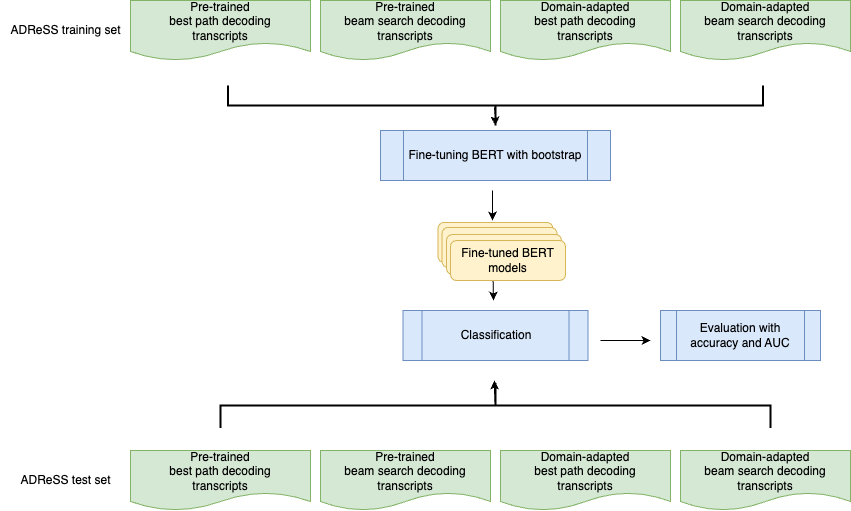}  
  \caption{Phase 2: fine-tuning BERT with participant-level ASR-generated transcripts to classify ASR-derived transcripts into those produced by dementia patients and healthy controls. The classification performance is evaluated by accuracy and AUC.}
  \label{fig:phase2}
\end{subfigure}
\caption{Overview of model development and evaluation for the downstream classification.}
\label{fig:flow}
\end{figure}

Figure~\ref{fig:flow} provides an overview of the study. The model development and evaluation consists of two phases: a) \textbf{transcript generation:} generating transcripts from utterance-level audio segments with pre-trained and domain-adapted ASR models, and b) \textbf{text classification:} fine-tuning BERT to classify ASR-derived transcripts into those produced by dementia patients and those from healthy controls. 

For the transcript generation phase, we forwarded utterance-level audio segments of ADReSS training and test sets to i) pre-trained ASR models, and ii) the same pre-trained ASR models adapted to the ''Cookie Theft'' picture description task on WLS training set with a batch size of 4 over 10 epochs, as the domain-adapted ASR models. Additionally, we experimented with using the ASR models in (i) and (ii) above in two distinct modes to generate transcripts: 1) using best-path decoding , which is the default generation method of all ASR models, and 2) using beam search decoding with a 5-gram LM trained on the remaining 16,043 ``perfect'' utterance-level WLS transcripts with the KenLM LM toolkit\footnote{\url{https://kheafield.com/code/kenlm/}} \citep{heafield-2011-kenlm}. 

We evaluated the performance of ASR models with word error rate (WER) and character error rate (CER) metrics calculated as shown in Equation~\ref{eq:wer}, where $S$ is the number of substitutions, $D$ is the number of deletions, $I$ is the number of insertions, $C$ is the number of correct words/characters, and $N$ is the number of words/characters in the reference.

\begin{equation}
    W/CER = \frac{S+D+I}{N} = \frac{S+D+I}{S+D+C}
    \label{eq:wer}
\end{equation}

For the classification task, we followed the previous state-of-the-art approach \citep{balagopalan20_interspeech} to establish baseline performance with the BERT model. We used bootstrapping in order to estimate the variability in model performance and establish confidence intervals around point estimates for classification accuracy and AUC. Specifically, we repeated the fine-tuning and evaluation process 100 times, then calculated the 95\% t-distribution confidence intervals for the resulting 100 pairs of accuracy and AUC metrics.

\subsection{Error Analysis}
We used SHapley Additive exPlanations (SHAP) \citep{NIPS2017_7062} to examine features influencing the classification decisions made by the fine-tuned BERT model with ASR-generated transcripts. Inspired by game theory, SHAP adapts the notion of Shapley values, which measure the contributions to the final outcome from each individual player among a coalition. When applied to BERT classification models, SHAP gives an overview of how each individual input feature (i.e., every ASR-generated character/token/word, in our case) contributes to the final classification prediction by computing Shapley values, which can be viewed as an expected value of predicted possibility for a given label. A positive Shapley value suggests a positive impact on the prediction, leading the fine-tuned BERT model to predict category ``1'' (i.e., dementia case). A negative Shapley value suggests a negative impact on a prediction of ``1''.

Furthermore, we examined \textit{all} transcripts generated by the adapted {wav2vec2-large-960h} model, which obtained the best subsequent classification performance with the beam search decoding method. We then calculated and summed Shapley values of short phrases similar to content units used in \citet{yorkston1980analysis} and compared their contributions to the final prediction. A ``content unit'' is defined here as one or more words (i.e., short phrases) that are always expressed as a unit by the speaker when describing an activity in the ``Cookie Theft'' picture \citep{yorkston1980analysis}. Note that short phrases selected by SHAP are usually not exact lexical matches to the corresponding content units as defined. To reduce variability, we grouped semantically similar short phrases with the corresponding content units for the purposes of the error analysis. We also treated phrases selected by SHAP that overlapped with a content unit by any amount as a match with the content unit concerned.

\section{Results}

\subsection{Transcript Generation Performance}

\begin{table}[htbp]
\centering
\begin{tabular}{|lll|}
\hline
\multicolumn{1}{|p{2.9cm}|}{\multirow{2}{*}{\textbf{Model}}} & \multicolumn{1}{l|}{\textbf{ADReSS test set}} &  \textbf{WLS test set}\\ \cline{2-3} 
\multicolumn{1}{|l|}{}   & \multicolumn{1}{l|}{WER (CER)} & WER (CER) \\ \hline
\multicolumn{3}{|c|}{\textit{Pre-trained, best-path decoding}} \\ \hline
\multicolumn{1}{|l|}{wav2vec2-base-960h} & \multicolumn{1}{l|}{0.559 (0.357)} &  0.541 (0.318)\\ \hline
\multicolumn{1}{|l|}{wav2vec2-large-960h} & \multicolumn{1}{l|}{0.493 (0.292)} & 0.471 (0.269) \\ \hline
\multicolumn{1}{|l|}{wav2vec2-large-960h-lv60} & \multicolumn{1}{l|}{0.443 (0.252)} & 0.412 (0.240)\\ \hline
\multicolumn{1}{|l|}{wav2vec2-large960h-lv60-self} & \multicolumn{1}{l|}{0.422 (0.258)} &0.390 (0.235) \\ \hline
\multicolumn{1}{|l|}{hubert-large-ls960-ft} & \multicolumn{1}{l|}{\textbf{0.415 (0.228)}} & \textbf{0.322 (0.210)} \\ \hline
\multicolumn{3}{|c|}{\textit{Domain-adapted, best-path decoding}}\\ \hline
\multicolumn{1}{|l|}{wav2vec2-base-960h} & \multicolumn{1}{l|}{0.438 (0.299)} &  0.366 (0.227)\\ \hline
\multicolumn{1}{|l|}{wav2vec2-large-960h} & \multicolumn{1}{l|}{0.427 (0.266)} & 0.358 (0.210) \\ \hline
\multicolumn{1}{|l|}{wav2vec2-large-960h-lv60} & \multicolumn{1}{l|}{0.364 (0.254)} & 0.277 (0.176)\\ \hline
\multicolumn{1}{|l|}{wav2vec2-large960h-lv60-self} & \multicolumn{1}{l|}{0.354 (0.234)} &0.264 (0.159) \\ \hline
\multicolumn{1}{|l|}{hubert-large-ls960-ft} & \multicolumn{1}{l|}{\textbf{0.332 (0.210)}} & \textbf{0.306 (0.153)} \\ \hline
\multicolumn{3}{|c|}{\textit{Pre-trained, beam search decoding}}\\ \hline
\multicolumn{1}{|l|}{wav2vec2-base-960h} & \multicolumn{1}{l|}{0.469 (0.335)} &  0.435 (0.293)\\ \hline
\multicolumn{1}{|l|}{wav2vec2-large-960h} & \multicolumn{1}{l|}{0.403 (0.267)} & 0.380 (0.243) \\ \hline
\multicolumn{1}{|l|}{wav2vec2-large-960h-lv60} & \multicolumn{1}{l|}{0.341 (0.223)} & 0.323 (0.215)\\ \hline
\multicolumn{1}{|l|}{wav2vec2-large960h-lv60-self} & \multicolumn{1}{l|}{0.340 (0.245)} &0.319 (0.224) \\ \hline
\multicolumn{1}{|l|}{hubert-large-ls960-ft} & \multicolumn{1}{l|}{\textbf{0.318 (0.225)}} & \textbf{0.305 (0.196)} \\ \hline
\multicolumn{3}{|c|}{\textit{Domain-adapted, beam search decoding}}\\ \hline
\multicolumn{1}{|l|}{wav2vec2-base-960h} & \multicolumn{1}{l|}{0.380 (0.290)} &  0.303 (0.220)\\ \hline
\multicolumn{1}{|l|}{wav2vec2-large-960h} & \multicolumn{1}{l|}{0.361 (0.253)} & 0.301 (0.205) \\ \hline
\multicolumn{1}{|l|}{wav2vec2-large-960h-lv60} & \multicolumn{1}{l|}{0.321 (0.250)} & 0.235 (0.174)\\ \hline
\multicolumn{1}{|l|}{wav2vec2-large960h-lv60-self} & \multicolumn{1}{l|}{0.310 (0.226)} &0.220 (0.154) \\ \hline
\multicolumn{1}{|l|}{hubert-large-ls960-ft} & \multicolumn{1}{l|}{\textbf{0.285 (0.205)}} & \textbf{0.210 (0.145)} \\ \hline
\end{tabular}
\caption{The performance of ASR models on the ADReSS and WLS test set. The best performance for each ASR model and generation method pair is in \textbf{bold}.}
\label{tab:asr-performance}
\end{table}

Table~\ref{tab:asr-performance} shows the ASR model performance using the best-path decoding method and beam search decoding on the ADReSS and WLS test set. While ASR accuracy with these dementia datasets falls far short of that with LibriSpeech data, both beam search decoding and domain adaptation (of the model itself on general-domain data, or of a post-processing language model using WLS data) improve performance considerably.

\subsection{Classification Performance}
\begin{figure*}
\small
\centering
\hskip-4cm\begin{minipage}{1.2\textwidth}
\begin{subfigure}{.6\textwidth}
  \centering
  \includegraphics[width=\linewidth]{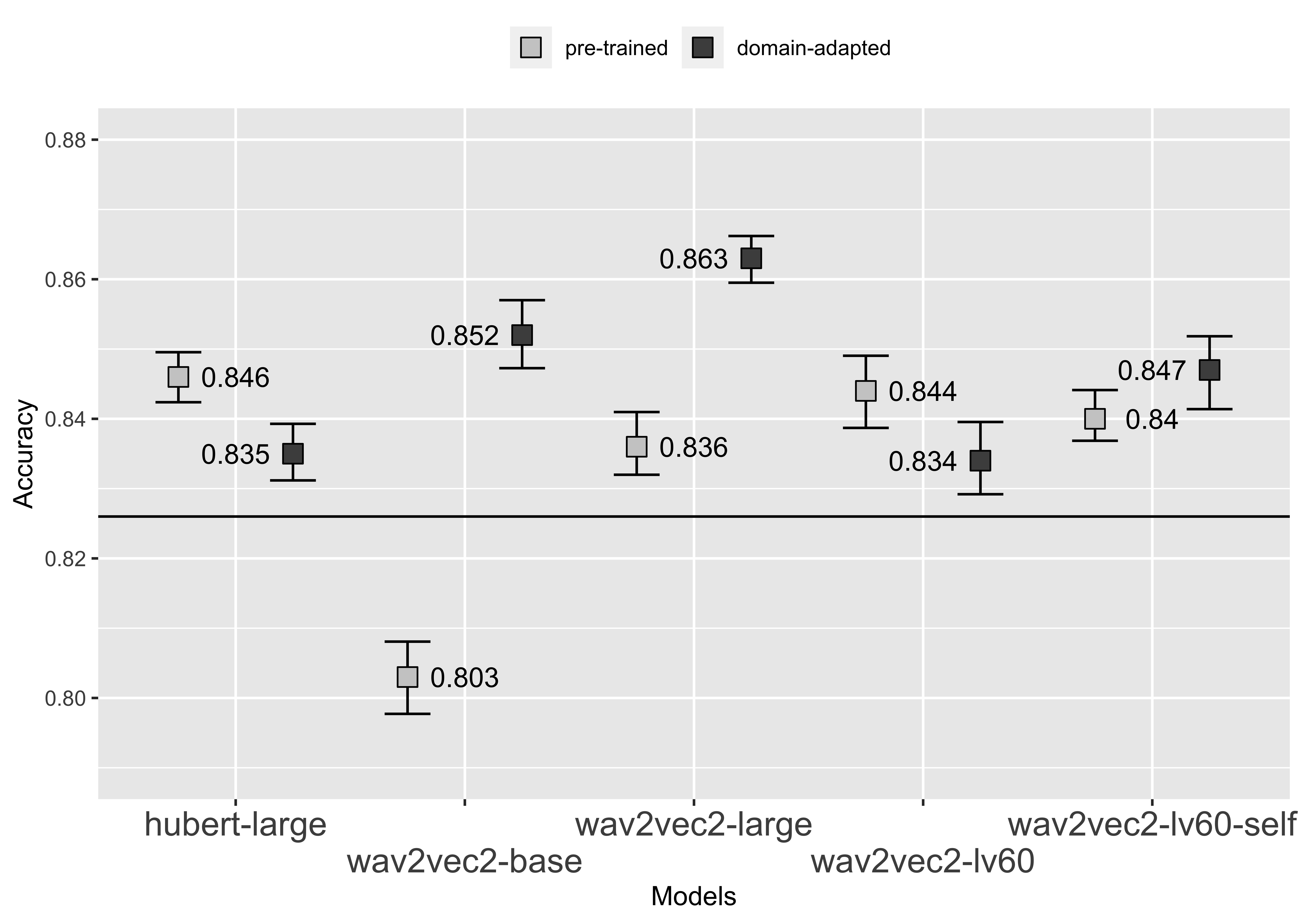}  
  \caption{ACC, with best-path decoding}
  \label{fig:trad-acc}
\end{subfigure}
\begin{subfigure}{.6\textwidth}
  \centering
  \includegraphics[width=\linewidth]{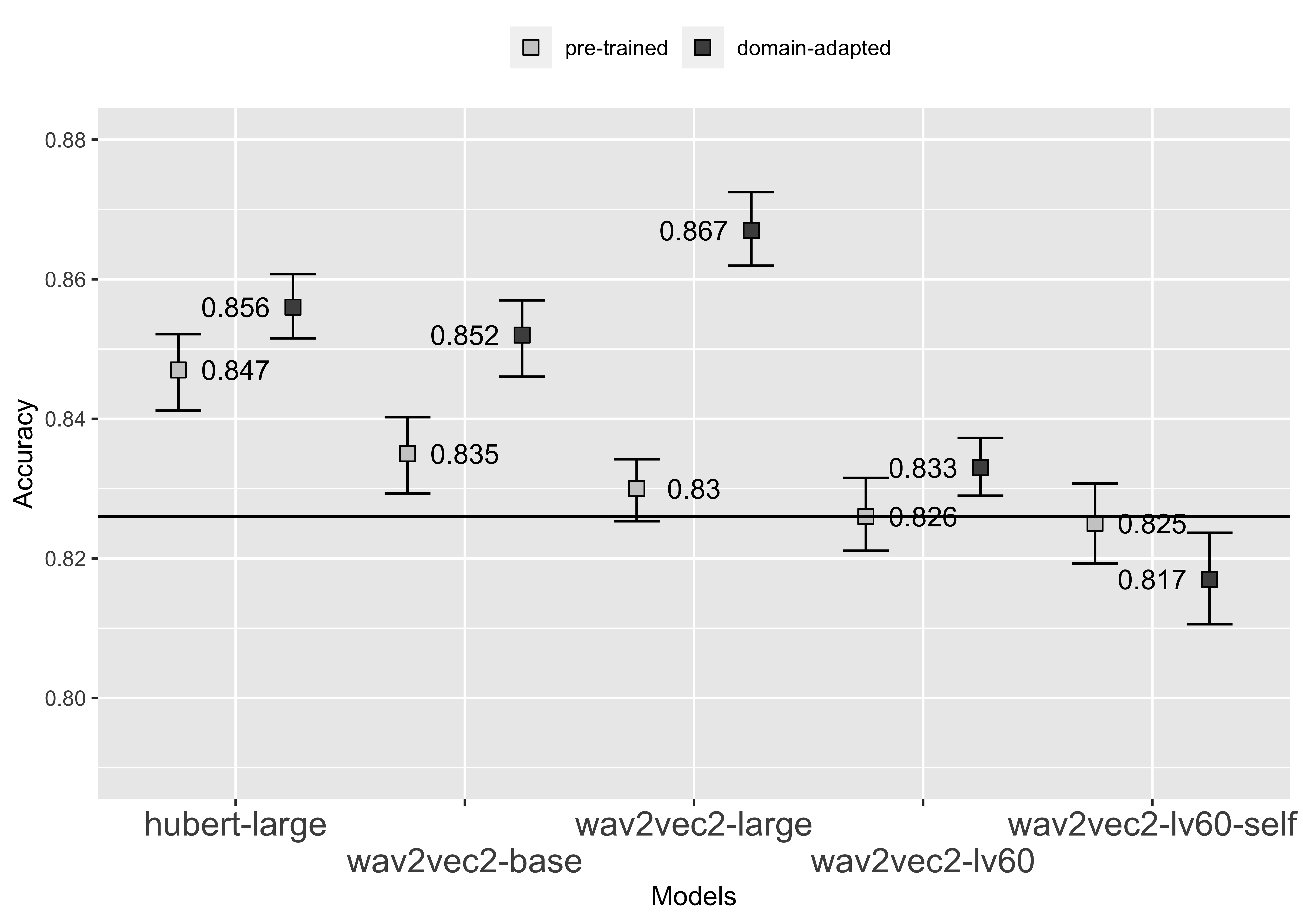}  
  \caption{ACC, with beam search decoding}
  \label{fig:ctc-acc}
\end{subfigure}

\begin{subfigure}{.6\textwidth}
  \centering
  \includegraphics[width=\linewidth]{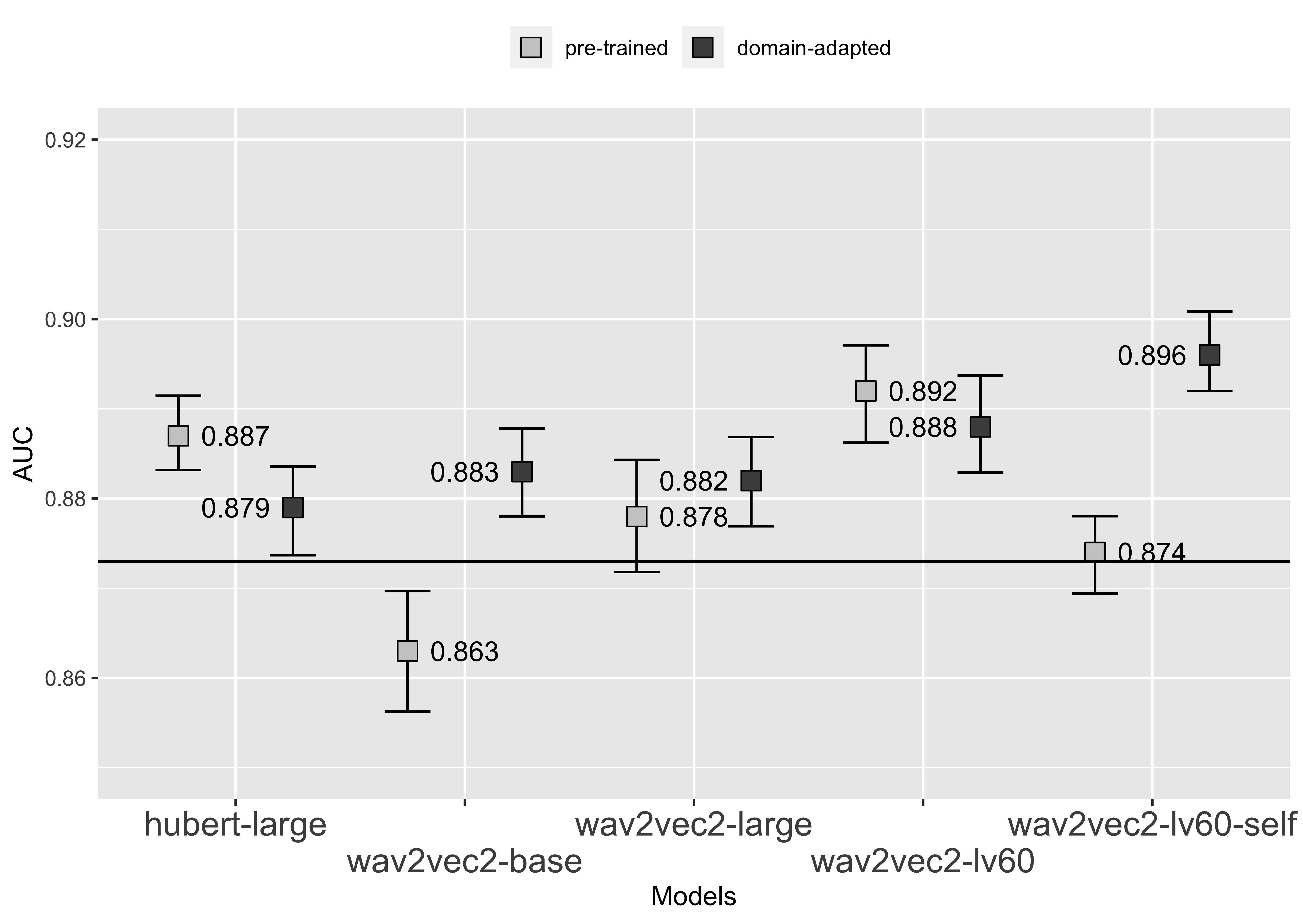}  
  \caption{AUC, with best-path decoding}
  \label{fig:trad-auc}
\end{subfigure}
\begin{subfigure}{.6\textwidth}
  \centering
  \includegraphics[width=\linewidth]{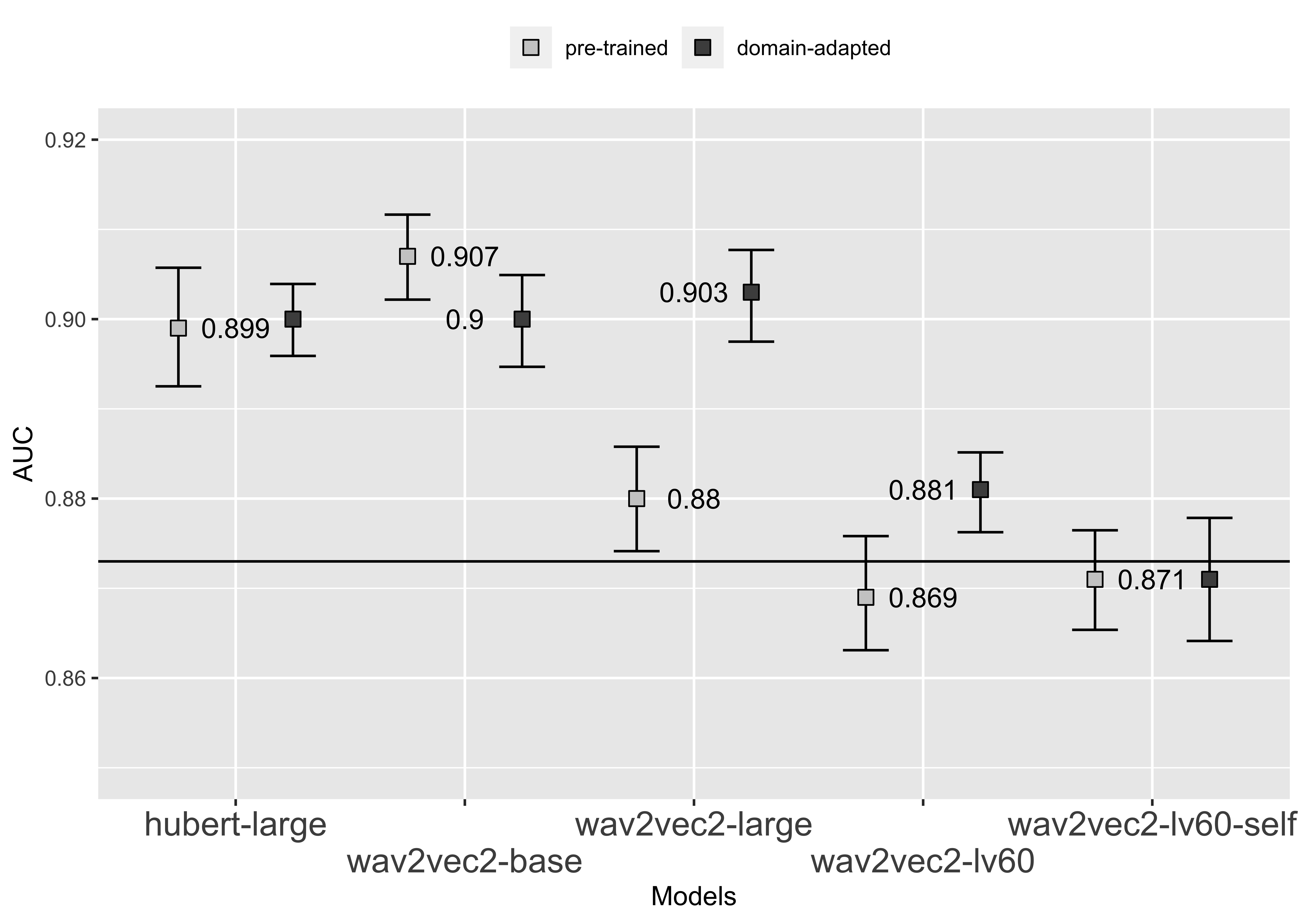}  
  \caption{AUC, with beam search decoding}
  \label{fig:ctc-auc}
\end{subfigure}
\caption{Classification performance, including accuracy (ACC) and AUC, and the corresponding 95\% t-distribution confidence interval with bootstrap. Metrics was calculated on participant-level ASR-generated transcripts. The horizontal lines represent the performance of the BERT model using manually-derived transcripts, with ACC of 0.826 and AUC of 0.873. wav2vec2-base, wav2vec2-large, wav2vec2-lv60, wav2vec2-lv60-self, and hubert-large represent wav2vec2-base-960h, wav2vec2-large-960h, wav2vec2-large-960h-lv60, wav2vec2-large-lv60-self, and hubert-large-ls960-ft, respectively.}
\label{fig:classification-performance}
\end{minipage}
\end{figure*}

As shown in Figure~\ref{fig:classification-performance}, surprisingly, performance on the subsequent classification task using imperfect transcripts generated by various ASR models generally exceeds the classification performance based on manually-derived verbatim transcripts (the horizontal line in each panel). However, when using the best-path decoding method, the downstream classification using transcripts generated by pre-trained {wav2vec2-base-960h} showed inferior performance - all 100 pairs of bootstrapped accuracy results fell behind those from the baseline model.

In addition to the comparison to the classification performance based on the manual transcripts (horizontal line), Figure~\ref{fig:classification-performance} provides a comparison between pre-trained (grey boxes) and domain-adapted after pre-training  (black boxes) variants of each model. In contrast to the evaluation of ASR accuracy, where domain adaptation improved performance across datasets, there is an inconsistent relationship between domain-adaptation and performance on the downstream classification task: sometimes the blue boxes are higher (indicating better performance), and sometimes the red boxes are higher. The best overall performance on the subsequent classification was achieved using transcripts generated by domain-adapted {wav2vec2-large-960h}, with accuracy at 0.863 and AUC at 0.882.

The downstream classification task did benefit from both the beam search decoding method and domain adaptation. In Figure~\ref{fig:ctc-acc} and Figure~\ref{fig:ctc-auc}, we observed that the performance of the classifier downstream of most of the ASR model variants outperformed the baseline. 

We also noticed that the downstream classification task using the transcripts generated from domain-adapted {wav2vec2-large-960h} with the beam search decoding method achieved the best performance in our study, achieving the accuracy of 0.867 and the AUC of 0.903 despite the relatively high WER/CER. Nevertheless, we noticed that the downstream task using transcripts generated from both pre-trained and domain-adapted {wav2vec2-large-960h-lv60-self} was inferior compared to the baseline model. 

\subsection{Error Analysis}

We further analyzed the errors made by the ASR models together with BERT predictions to investigate how the erroneous ASR-generated transcripts could lead to subsequent classification performance that is equivalent to, or even better than, that with manual verbatim transcripts. We selected three examples from the ADReSS test set from participants with different levels of cognitive impairment to perform the error analysis: a) ADReSS ID 114-1, with MMSE score of 30, labeled as healthy control, and correctly classified by baseline BERT as healthy control; b) ADReSS ID 148-1, with MMSE score of 10, labeled as dementia case, but incorrectly classified by the baseline BERT as healthy control, and c) ADReSS ID 150-2, with MMSE score of 24, and correctly classified by the baseline BERT model as healthy control.

\subsubsection{hubert-large-ls960-ft}
\begin{sidewaysfigure*}[htbp]
\small
\centering
\hskip-3cm\begin{minipage}{1.0\textwidth}
\begin{subfigure}{.5\textwidth}
  \centering
  \includegraphics[width=\linewidth]{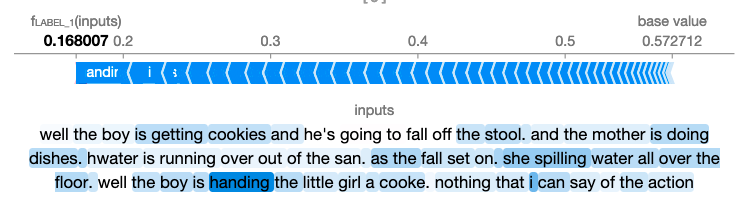}  
  \caption{SHAP values for ADReSS ID 114-1 (healthy control) with pre-trained {hubert-large-ls960h-ft}}
  \label{fig:ori-hubert-control}
\end{subfigure}\hskip1cm
\begin{subfigure}{.5\textwidth}
  \centering
  \includegraphics[width=\linewidth]{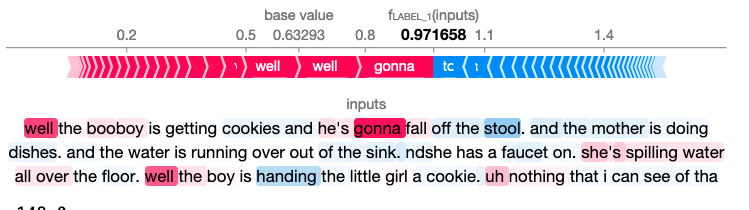}  
  \caption{SHAP values for ADReSS ID 114-1 (healthy control) with domain-adapted {hubert-large-ls960h-ft}}
  \label{fig:ft-hubert-control}
\end{subfigure}
\begin{subfigure}{.5\textwidth}
  \centering
  \includegraphics[width=\linewidth]{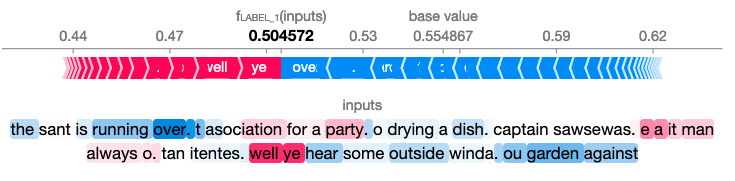}  
  \caption{SHAP values for ADReSS ID 148-0 (dementia) with pre-trained {hubert-large-ls-960h}}
  \label{fig:ori-hubert-dementia}
\end{subfigure}\hskip1cm
\begin{subfigure}{.5\textwidth}
  \centering
  \includegraphics[width=\linewidth]{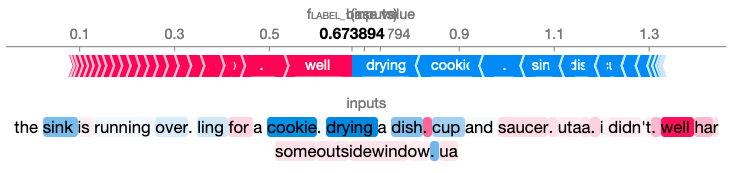}  
  \caption{SHAP values for ADReSS ID 148-0 (dementia) with domain-adapted {hubert-large-ls960h}}
  \label{fig:ft-hubert-dementia}
\end{subfigure}
\begin{subfigure}{.5\textwidth}
  \centering
  \includegraphics[width=\linewidth]{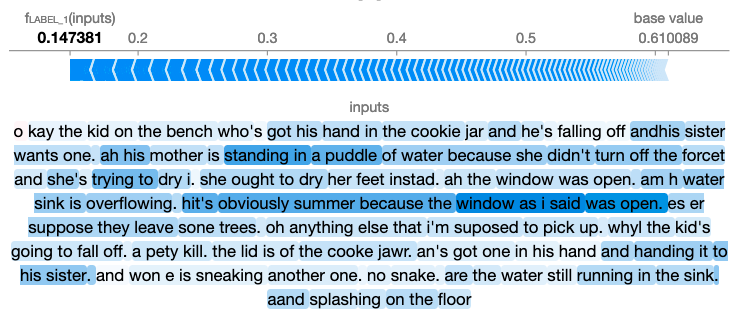}  
  \caption{SHAP values for ADReSS ID 150-2 (healthy control) with pre-trained {hubert-large-ls960h-ft}}
  \label{fig:ori-hubert-mci}
\end{subfigure}\hskip1cm
\begin{subfigure}{.5\textwidth}
  \centering
  \includegraphics[width=\linewidth]{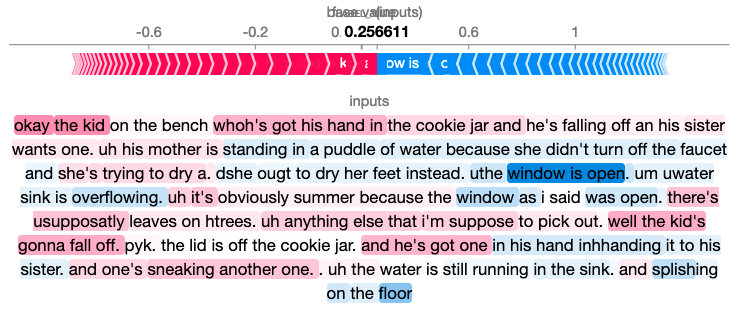}  
  \caption{SHAP values for ADReSS ID 150-2 (healthy control) with domain-adapted {hubert-large-ls960h}}
  \label{fig:ft-hubert-mci}
\end{subfigure}
\caption{The visual representation of Shapley values for ADReSS ID 114-1 (healthy control), 148-0 (dementia), and 150-2 (healthy control), using pre-trained and domain-adapted {hubert-large-ls960h-ft} with the best-path decoding. The $f_{LABEL\_1}(inputs)$ represents the expected possibility of this transcript being produced by a dementia patient. Tokens in red represent positive impact on the prediction, whereas tokens in blue represent negative impact on the prediction. Darker color represents higher degree of the corresponding impact.}
\label{fig:shap-hubert}
\end{minipage}
\end{sidewaysfigure*}

\begin{sidewaysfigure*}[htbp]
\small
\centering
\hskip-3cm\begin{minipage}{1.0\textwidth}
\begin{subfigure}{.5\textwidth}
  \centering
  \includegraphics[width=\linewidth]{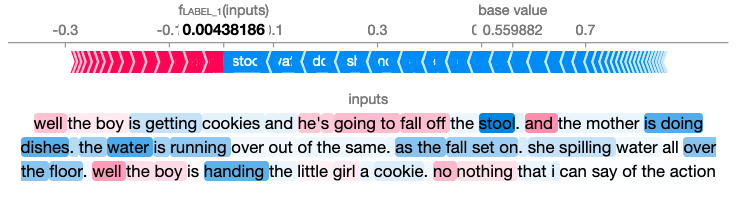}  
  \caption{SHAP values for ADReSS ID 114-1 (healthy control) with pre-trained {hubert-large-ls960h-ft}}
  \label{fig:ori-hubert-control-ctc}
\end{subfigure}\hskip1cm
\begin{subfigure}{.5\textwidth}
  \centering
  \includegraphics[width=\linewidth]{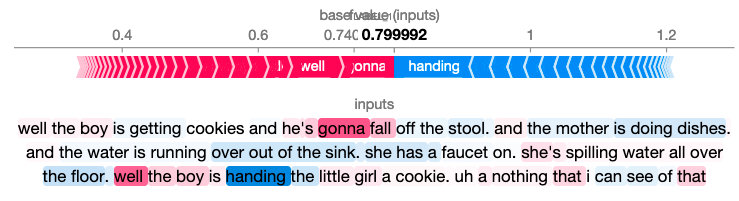}  
  \caption{SHAP values for ADReSS ID 114-1 (healthy control) with domain-adapted {hubert-large-ls960h-ft}}
  \label{fig:ft-hubert-control-ctc}
\end{subfigure}
\begin{subfigure}{.5\textwidth}
  \centering
  \includegraphics[width=\linewidth]{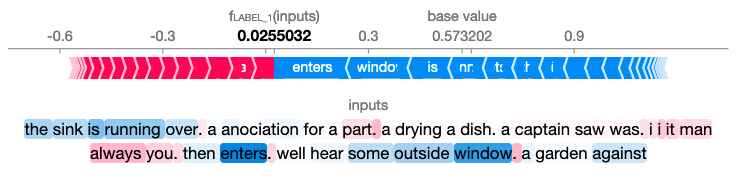}  
  \caption{SHAP values for ADReSS ID 148-0 (dementia) with pre-trained {hubert-large-ls-960h}}
  \label{fig:ori-hubert-dementia-ctc}
\end{subfigure}\hskip1cm
\begin{subfigure}{.5\textwidth}
  \centering
  \includegraphics[width=\linewidth]{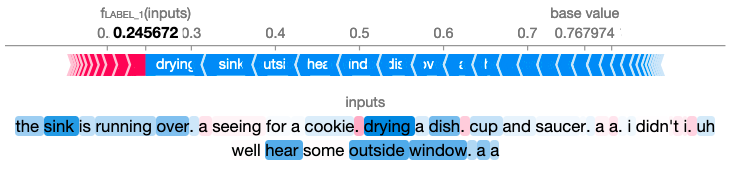}  
  \caption{SHAP values for ADReSS ID 148-0 (dementia) with domain-adapted {hubert-large-ls960h}}
  \label{fig:ft-hubert-dementia-ctc}
\end{subfigure}
\begin{subfigure}{.5\textwidth}
  \centering
  \includegraphics[width=\linewidth]{ori-hubert-mci.png}  
  \caption{SHAP values for ADReSS ID 150-2 (healthy control) with pre-trained {hubert-large-ls960h-ft}}
  \label{fig:ori-hubert-mci-ctc}
\end{subfigure}\hskip1cm
\begin{subfigure}{.5\textwidth}
  \centering
  \includegraphics[width=\linewidth]{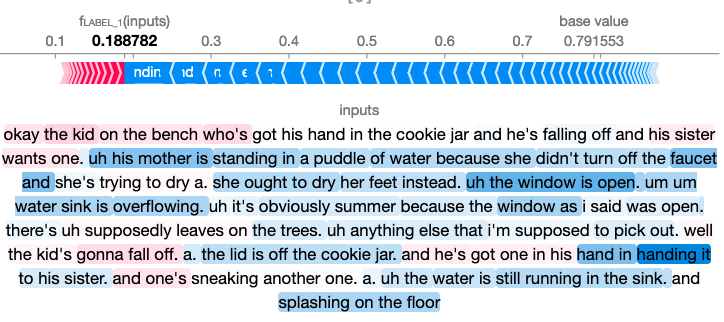}  
  \caption{SHAP values for ADReSS ID 150-2 (healthy control) with domain-adapted {hubert-large-ls960h}}
  \label{fig:ft-hubert-mci-ctc}
\end{subfigure}
\caption{The visual representation of Shapley values for ADReSS ID 114-1 (healthy control), 148-0 (dementia), and 150-2 (healthy control), using pre-trained and domain-adapted {hubert-large-ls960h-ft} with the beam search decoding. The $f_{LABEL\_1}(inputs)$ represents the expected probability of this transcript being produced by a dementia patient. Tokens in red represent positive impact on the prediction, whereas tokens in blue represent negative impact on the prediction. Darker color represents higher degree of the corresponding impact.}
\label{fig:shap-hubert-ctc}
\end{minipage}
\end{sidewaysfigure*}
The best-performing ASR model, pre-trained {hubert-large-ls960h-ft}, obtained the lowest WERs and CERs. In the subsequent classification task using transcripts generated from this model, BERT correctly classified all three data samples. However, with additional domain adaptation, the subsequent classification model mistakenly classified 114-1 as a dementia case. 
We can observe that words related to the content of the ``Cookie Theft'' picture contribute more to the prediction of healthy controls while the words related to the children's activities, non-words, and misspelled words contribute more to the prediction of dementia.

The beam search decoding further improved the ASR performance of the pre-trained {hubert-large-ls960-ft} model by generating more accurate transcripts. However, such improvements did not benefit the downstream classification task. For example, pre-trained {hubert-large-ls960-ft} with the beam search decoding method generated \texttt{the sink is running over. a anociation for a part. a drying a dish. a caption saw was. i. i it man always you. then enters. well hear some outside window. a garden against} for 148-0 (Figure~\ref{fig:ori-hubert-dementia-ctc}). However, the additional context brought by the LM decreased the predicted probability of being a dementia case, which dropped from 0.505 (Figure~\ref{fig:ori-hubert-control}) to 0.026. Additional domain adaptation with the beam search decoding method improved the accuracy of transcription. However, it also degraded the downstream classification task. As shown in Figure~\ref{fig:shap-hubert-ctc}, the downstream classification model mistakenly classified 114-1 as a dementia case and 148-0 as a healthy control.

\subsubsection{Content Units}
\begin{table*}[htbp]
    \centering
    \tiny
    \hskip-3cm\begin{minipage}{0.8\textwidth}\begin{tabular}{|p{3cm}|p{9.5cm}|l|}
    \hline
        \textbf{Verbatim content units} & \textbf{ASR-generated content units} & \textbf{Values} \\ \hline
        two & the two/two children & 0.024 \\ \hline
        children & kids/children/kid/children evidently/child & 0.481 \\ \hline
        little & little/the little/a little/his little & 0.958 \\ \hline
        boy & a dish and the/dishes/the dishes/dish/she has a dish/plate/dish looking/the dishe that's/a good boy to & 0.622 \\ \hline
        brother & brother & 0.024 \\ \hline
        standing & standing/standing in the/standing on a/boy standing/is standing/standing on a so/standing on/the boy is standing & -0.24 \\ \hline
        on stool & stool/chair that/stool which/ladder/the stool/a time on stool/tipped stool there/on the stool/a stool which is/here the boy that's got ladder/in his chair/chair/off the stool & -1.212 \\ \hline
        wobbling (off balance) & slip from/is bending over/trunning over/balance/slipping out & -0.246 \\ \hline
        3-legged & um three/a three legs/three legged stool & -0.098 \\ \hline
        falling over & is tippng over/falling off/flowing out/falling off the stool/stool falling out over/falling/is starting to fall over/off/the boy is falling off the/is about to fall & -0.326 \\ \hline
        on the floor & on the floor & -0.068 \\ \hline
        hurt himself & break his neck & -0.041 \\ \hline
        reaching up & grabbing/getting/is reaching into the/reaching for a cookie/reaching up/reaching/there reaching/reaching in/grabbing for/the reaching/up reach /reaching at/he's reaching for one for himself & -0.732 \\ \hline
        taking (stealing) & steaking/stool stealing/thing the boy is stealing cookies/are in the process of stealing cookies from/and he's stealing cookies out of/they're steal  & 0.068 \\ \hline
        cookie & cookie jar/the cookies/a cookie/the cookie jar he/pookies/out of the cookie jar/the cookies out of/for cookies/of a cookie jar/cookie in/some cookie cookies out/you gonna cookies & 0.321 \\ \hline
        for himself & from his hand & -0.099 \\ \hline
        in the cupboard & cupboards/cupborad in/shelf of the cupbaord & -0.32 \\ \hline
        with the open door & ~ & ~ \\ \hline
        handing to sister & handing the little girl a cookie/he's handing the little girl/handing one down to/hand it to his sister & -0.191 \\ \hline
        girl & girl/the girl/the little girl is/the girl cookie out/a nicely a girl with a/the little girl is/girl a talking/the little has/his sister & 0.1308 \\ \hline
        asking for cookie & be as she wants/bedding for cookies & 0.046 \\ \hline
        has finger to mouth & going mouth/is touching her lips/putting her finger to her mouth that she's/the finger to her lips/putting her finger tool or lips/her finger mount/her finger up to & -0.19 \\ \hline
        saying shhh (keeping him quiet) & to keep it quiet/whispering/quite/her mom be quiet/quite & -0.435 \\ \hline
        laughing & laughing at here/laughing & -0.042 \\ \hline
        mother & mother washing dishes/mother's/the mother/mother is/woman's mother is/mom's/mother might/ther're mom & -0.025 \\ \hline
        woman (lady) & the lady's/woman's & -0.145 \\ \hline
        children behind her & ~ & ~ \\ \hline
        by sink & sink is/the sink/on the sink/running in the in the sink/out of the sink/and the sink/sink/in the sink and/watching sink/in the kitchen sink/of the sink/a sink & -0.214 \\ \hline
        washing (doing) & is washing dishes/washing dises/washing a for/they're washing/washing the dishes/wash to/doing/spilling the woman washing the dishes & 0 \\ \hline
        dishes & a dish and the/dishes/the dishes/dish/she has a dish/plate/dish looking/the dishe that's & -0.309 \\ \hline
        drying & apron drying dishes/drying/drying the/drying in/drying to/dry the/drying the dishes/she's drying dishes/dromg a dish & -0.611 \\ \hline
        faucet on & faucet/and a saucer/faucet running & -0.044 \\ \hline
        ignoring (daydreaming) & aware/daydreaming & -0.058 \\ \hline
        water & water's/water/running the water on/and the water is running/water coming/the water/then spilling the water/with water/tippling water on the/ih water sink & -0.432 \\ \hline
        overflowing & overflowing/overflow/running out/is running over/is running/flowing running flowing out/is running over/over out/the sink is running over/tipping over/spilling over & -1.832 \\ \hline
        onto floor & on the floor/the floor/down onto the floor/all over the floor/on to the floor & -0.4 \\ \hline
        feet getting wet & getting her feet wet/standing in the water/she on to dry her feet and & -0.21 \\ \hline
        puddle & puddle & -0.036 \\ \hline
        in the kitchen (indoors) & kitchen/the kitchen/if the kitchen who & -0.006 \\ \hline
        open window & window's got open/window/window's open/the window/the window is open/out the window/outside window/the window is open/window as was open & -0.811 \\ \hline
        curtains & curtain/the curtains/the curtains seem to/on the kitchen curtains/and the curtains & -0.627 \\ \hline
        \multicolumn{3}{|p{1.2\textwidth}|}{Content units not found in ASR transcripts: disaster, lawn, sidewalk, nextdoor house, full blast, dirty dishes left, children behind her, trying to helo (or not), by boy, from the jar, on the high shelf}  \\ \hline
    \end{tabular}
    \end{minipage}
    \caption{The comparison of content units between verbatim and ASR-generated transcripts. Tokens with positive Shapley values contribute more to a dementia case prediction whereas ones with negative Shapley values contribute more to a healthy control prediction. Note that the verbatim content units cannot be directly applied to the ASR-generated transcripts, as tokenization is different from two methods. The value column indicates the summed Shapley values for all ASR-generated content units.}
    \label{tab:content_units}
\end{table*}

Table~\ref{tab:content_units} shows the comparison of content units between verbatim and ASR-generated transcripts. We observed that content units related to the children (i.e., ``children'', ``boy'', ``brother'', ``little'', and ``girl'') contributed more to dementia case predictions, resulting in positive Shapley values. Interestingly, when participants referred to the mother (i.e., ``woman/lady''), such content units contributed more to healthy control predictions, resulting in negative Shapley values. However, when participants referred to the woman as ``mother/mom'' or ``she'', it resulted in positive Shapley values, which indicates that such content units contributed more to dementia case predictions, in contrast to the content units such as ``woman/lady''. We also observed that content units referring to the activity of children (i.e., ``standing'', ``on stool'', ``wobbling/off balance'', and ``reaching up'') contributed more to healthy control predictions. Surprisingly, content units referring to the activity of stealing cookie (i.e. ``talking/stealing'', ''cookie'', and ``asking for cookie'') contributed more to dementia case predictions. Content units related to the mother's activity (i.e., ``washing'', ``dishes'', ``drying'') and the condition of the sink (i.e., ``water'', ``overflowing'', ``onto floor'') contributed significantly to healthy control predictions.

When participants referred to the environment of the ``Cookie Theft'' picture, the related content units (i.e., ``open window'', ``curtains'', ``feet getting wet'' and ``faucet on'') significantly contributed to healthy control prediction with negative Shapley values. Additionally, we observed that when participants repeated themselves, the corresponding content units usually got positive Shapley values, indicating greater contribution to dementia case predictions. Along with content units, we found that non-content units such as ``well'', ``I can'', ``that's it'', ``that's'' and ``I don't think'' often contributed significantly to dementia case prediction with positive Shapley values. When ASR models cannot pick up any useful acoustic signals thus generating phonetic-level transcripts (i.e., ``i'', ``a'') or empty transcripts, those transcripts often got positive Shapley values as well.

\section{Discussion}

The findings in this work are consistent with our previous study \citep{li2022far}: ASR transcripts with high WERs and CERs (i.e., worse ASR accuracy) seem to provide important features for the downstream classification task, resulting in better performance in many cases than models trained and tested on manually created verbatim transcripts. In particular, the imperfect ASR-generated transcripts from both pre-trained and domain-adapted ASR models generally outperform the baseline model trained using manual transcripts. Furthermore, when introducing the beam search decoding method, the ASR-generated transcripts were found to be closer to manual verbatim transcripts. Surprisingly, these improvements did not benefit performance on the downstream classification task. As shown in our error analysis, the downstream classification model tended to recognize those correctly generated phrases as patterns of healthy controls.

\subsection{Effects of improvements in ASR accuracy on downstream classification}

As we expected, domain adaptation improved the performance of ASR models in terms of the WER/CER characteristics. Our results are consistent with another previous study \citep{min-etal-2021-evaluating} showing that adapting on additional data from the domain may alleviate the impact of noisy transcriptions and audio recordings. Our findings are also consistent with prior observations that domain adaptation can be beneficial for downstream tasks \citep{9688093}. The observation of HuBERT's better performance is also consistent with prior work \citep{9688137} indicating that HuBERT supports the state-of-the-art performance in many downstream tasks, and can generalize well to multiple ``out-of-distrubtion'' datasets. Our qualitative observations of the differences in ASR transcripts suggest that HuBERT may have a better acoustic representation and is more robust to noise, which allows it to be more sensitive to the acoustic characteristics associated with cognitive impairment and ``express" them as nonsensical character sequences for audio which other models simply would filter out as noise. 

Combining domain adaptation and the beam search decoding methods further improved the performance of ASR models, however, these improvements restricted the predictive power of the downstream classification model. We believe that these somewhat surprising results indicate that there is a synergistic relationship between ASR models and downstream classification models in that the ASR models may react to dementia manifestations in both linguistic and acoustic contexts in systematic ways (manifesting as ASR errors) that may not be immediately apparent to human eye (or ear) but can be leveraged as useful features by neural classification models. For example, with training on ASR-derived transcripts, downstream classification models tend to treat ``irregular'' linguistic patterns (e.g., ``gonna'') and short phrases with non-words, single syllables and incorrectly spelled words due to ASR errors as dementia manifestations.

Interestingly, while pre-trained {wav2vec2-base-960h} achieved the highest WER at 0.559 and CER at 0.357 on the ADReSS set, the BERT classification that used the transcripts generated by this ASR model was only $\approx$ 2\% in both accuracy and AUC below the baseline state-of-the-art results obtained with ideal manual transcripts. On the other hand, the best performing ASR model, domain-adapted {hubert-large-ls960-ft} fused with an n-gram language model achieved the lowest WER at 0.285 and CER at 0.205 on the ADReSS test set. However, this decrease of $\approx$ 30\%  in WER and $\approx$ 20\% in CER only brings in $\approx$ 4\% increase in both accuracy and AUC. These results indicate that while there appears to be a generally positive relationship between ASR accuracy and downstream classification performance, this relationship may not always hold and may have a threshold, potentially necessitating an ASR error management system in support of this specialized downstream task \citep{balagopalan-etal-2020-impact}. Although it is unknown whether the ASR system's blunders contain information that is clinically significant, they appear to be advantageous for the downstream task. This paradoxical relationship needs to be further explored in future work but it clearly indicates that the errors produced by ASR models could also be useful in a clinically-related setting. 

We should also point out that the audio quality of ADReSS dataset is relatively low. The low quality contributed to degraded ASR model performance, as these models could detect less useful acoustic signals to generate meaningful transcripts. For example, we find that all transcripts having phonetic output (i.e., single characters such as ``i'', ``a'') lead to positive Shapley values, which contribute more to a dementia case prediction. We manually checked the audio segments and found that most of them were nearly inaudible - very quiet speech with a high level of channel noise. This suggests that the model may implicitly encode paralinguistic indicators of dementia or environmental factors (e.g., speaking away from the microphone) that may also be non-coincidental, which may in turn impact downstream classification accuracy. On the other hand, we noticed that ASR models, especially domain-adapted {hubert-large-ls960-ft}, obtain remarkable performance on the WLS dataset, reaching WER at 0.210 and CER at 0.145, suggesting that high quality ``Cookie Theft'' picture description recordings may provide better insights into the relationship between ASR errors and downstream classification performance by allowing better control of some of these potentially random factors.

\subsection{Improved interpretability of dementia classification decisions}
Similarly to other prior work using transcripts or audio signals from the ADReSS/WLS dataset for classification, our approach contributes improved interpretability to NLMs by providing transcripts generated by SSL ASR models and detailed error analysis of the downstream prediction, potentially offering more transparency for healthcare providers to understand the decisions made by DNNs. It is important to improve interpretability of NLP/ASR models (in addition to achieving the state-of-the-art performance) in the clinical/medical domain \citep{tjoa2020survey, ghassemi2021false, Holzinger2017WhatDW}. With improved interpretability, such models can be one step closer to becoming accurate and trustworthy tools for clinicians to screen and monitor progress of dementia, even without full comprehension of AI. With improved interpretability, NLP/ASR models can also help clinical decision-making and make it simpler for clinicians to communicate their findings to patients, their families, and fellow clinicians. 

We further extend the notion of content units by looking at the context of predicted labels from the downstream classification model. Unlike previous studies \citep{yorkston1980analysis, giles1996performance, nicholas1985empty} that focus on the language itself (i.e., semantic content, syntactic complexity, speech fluency, vocal parameters, and pragmatic language), our results show that the context of content units (i.e., activity related to children/mother, general description of people in the picture) also contribute toward the discriminating power of the downstream BERT classification model for differentiating between language produced by cognitively healthy controls and that from AD patients. The finding that repeated content units contribute more to a dementia case prediction is consistent with prior work on perseveration in dementia \citep{HIER1985117,ralph2001semantic}, suggesting that dementia patients' speech tends to be more repetitive than that of healthy controls. Furthermore, our observation of different Shapley values between ``woman/lady'' and ``mom/she'' may suggest the downstream classification model resembles impaired reasoning ability of dementia patients. Clearly, these observations are based on only a few examples and will need to be further explored in future work to determine if the context of content units may have an underlying structure with diagnostic utility.

\subsection{Limitations}

Our work has several limitations. First, the poor quality of ADReSS audio recordings is very challenging for ASR models, which prevented us from exploring the lower range of WER/CER in an ecologically valid fashion (i.e., without artificially manipulating ASR-generated transcripts). There is a critical need for public availability of high-quality audio recordings of picture description tasks, using the ``Cookie Theft" picture and other stimuli, performed by patients with dementia and controls and accompanied by neuropsychological evaluation scores. Second, the ADReSS and the WLS data are in American English, and many participants of these two studies are representative of White, non-Hispanic American men and women located at the north part of the United States. Performance of ASR models may differ in populations with different accents \citep{prasad-jyothi-2020-accents}. Third, it should be noted that all of the ASR models we tested in this study were pre-trained on read speech, whose nature is very different from the spontaneous speech \citep{8268245}. This may help to explain why WER and CER differ on the ``Cookie Theft" audio recordings, regardless of the audio quality. ASR models that are pre-trained on spontaneous speech would potentially benefit the downstream task. While NLMs can identify language anomalies caused by multiple brain mechanism and, while they can contribute to the screening and monitoring of the disease, they should not be considered as accurate and comprehensive representations of human cognition. It is important to note that the use of a relatively small dataset in this study may pose a potential overfitting concern, thus potentially constraining the generalizability of our findings.

\section{Conclusion}
In this work, we investigated the utility of using ASR-generated transcripts in lieu of manual transcripts for discriminating spontaneous speech of dementia patients from that of controls. We found that domain-adapting on a held-out dataset and fusing with an n-gram LM improved the ASR transcription accuracy. Surprisingly, the downstream neural classification model using error-prone transcripts as input outperformed the one using manually created verbatim transcripts. Worse ASR accuracy does not necessarily lead to worse subsequent classification performance and can even enhance performance, perhaps by allowing models to recognize ASR errors that occur systematically in the presence of impaired speech in dementia. Further work is needed to confirm these findings in larger and more diverse datasets. However, if confirmed, our findings would have significant implications for use of ASR technology to automate the analysis of speech collected from patients at least in the dementia screening settings but possibly in a variety of other clinical applications as well in which both language and speech characteristics are affected. Further work will be needed to understand the exact nature of the relationship between ASR errors and linguistic features of speech collected from patients with dementia, specifically, it will be important to understand if ASR models can by systematically manipulated to make them more sensitive to known linguistic manifestations of dementia.

\section*{Acknowledgement}
This research was supported by grants from the National Institute on Aging, AG069792, R01LM014056-02S1, and R21AG069792-01. 

\bibliographystyle{elsarticle-num-names}

\bibliography{cas-refs}

\end{document}